\documentclass[letterpaper, 10 pt, conference]{ieeeconf}  

\IEEEoverridecommandlockouts                              

\title{\LARGE \bf
Decentralized Social Navigation with Non-Cooperative Robots via Bi-Level Optimization
}

\author{Rohan Chandra, Rahul Menon, Zayne Sprague, Arya Anantula, and Joydeep Biswas \\ \small{\texttt{\{rchandra, rmeno12, zaynesprague, anantula.arya, joydeepb\}@utexas.edu}}\\\small{University of Texas, Austin}\\\small{Project hosted at \href{https://amrl.cs.utexas.edu/snupi.html}{\textbf{https://amrl.cs.utexas.edu/snupi.html}}}
\thanks{All authors are with the Department of Computer Science, University of Texas, Austin.}
}

\usepackage{color}
\usepackage{colortbl}
\usepackage{xcolor}
\newcolumntype{a}{>{\columncolor{green}}c}
\newcolumntype{b}{>{\columncolor{green!65}}c}
\newcolumntype{d}{>{\columncolor{green!40}}c}
\newcolumntype{e}{>{\columncolor{green!15}}c}
\usepackage{xspace}
\usepackage{multicol,multirow}

\usepackage{amsthm}
\usepackage{soul}
\usepackage{threeparttable}
\usepackage{pifont}
\usepackage{wrapfig}
\usepackage{amsmath, amsfonts, amssymb}   
\usepackage{mathtools}
\usepackage{pifont}%
\usepackage{booktabs}
\usepackage{float}
\usepackage[linesnumbered,ruled,vlined]{algorithm2e}
\usepackage{bm}
\usepackage{subcaption}
\usepackage[font=small]{caption}
\usepackage[colorlinks=true, citecolor=magenta]{hyperref}

\newtheorem{definition}{Definition}

\newtheorem{problem}{Problem}

\begin{document}
\maketitle
\thispagestyle{empty}
\pagestyle{empty}
\begin{abstract}
This paper presents a fully decentralized approach for realtime non-cooperative multi-robot navigation in social mini-games, such as navigating through a narrow doorway or negotiating right of way at a corridor intersection. Our contribution is a new realtime bi-level optimization algorithm, in which the top-level optimization consists of computing a fair and collision-free ordering followed by the bottom-level optimization which plans optimal trajectories conditioned on the ordering. We show that, given such a priority order, we can impose simple kinodynamic constraints on each robot that are sufficient for it to plan collision-free trajectories with minimal deviation from their preferred velocities, similar to how humans navigate in these scenarios.

We successfully deploy the proposed algorithm in the real world using F$1/10$ robots, a Clearpath Jackal, and a Boston Dynamics Spot as well as in simulation using the SocialGym 2.0 multi-agent social navigation simulator, in the doorway and corridor intersection scenarios. We compare with state-of-the-art social navigation methods using multi-agent reinforcement learning, collision avoidance algorithms, and crowd simulation models. We show that $(i)$ classical navigation performs $44\%$ better than the state-of-the-art learning-based social navigation algorithms, $(ii)$ without a scheduling protocol, our approach results in collisions in social mini-games $(iii)$ our approach yields $2\times$ and $5\times$ fewer velocity changes than CADRL in doorways and intersections, and finally $(iv)$ bi-level navigation in doorways at a flow rate of $2.8 - 3.3$ (ms)$^{-1}$ is comparable to flow rate in human navigation at a flow rate of $4$ (ms)$^{-1}$.
\end{abstract}
\section{Introduction}
\label{sec: introduction}

Navigating through a narrow door, passing in a narrow hallway, or negotiating right of way at a corridor intersection\footnote{We refer to such scenarios as \textit{``social mini-games''}} pose several challenges for socially compliant multi-robot navigation systems; two key challenges, in particular, demand attention. First, robots may be non-cooperative, seeking to optimize their own objective. For instance, robots with selfish motives arriving at a corridor intersection would prefer to move through the intersection first. This leads to behaviors that result in deadlocks, collisions, or jerkier and longer trajectories. Second, humans tend to exhibit \textit{socially compliant} trajectories that are characterized by minimal deviations from their preferred velocities. For instance, when two individuals go through a doorway together, one person modulates their velocity by a slight amount, just enough to enable the other person to pass through first, while still adhering closely to their preferred speed. This type of behavior presents a significant challenge for robots, which struggle to emulate such socially adaptive maneuvers while maintaining a consistent preferred velocity. The objective of this paper is to propose a new approach for safe, efficient, and socially-compliant navigation for multiple non-cooperative robots in social mini-games.

\begin{figure}[t]
    \centering
    \begin{subfigure}[h]{0.485\columnwidth}
\includegraphics[width = \columnwidth]{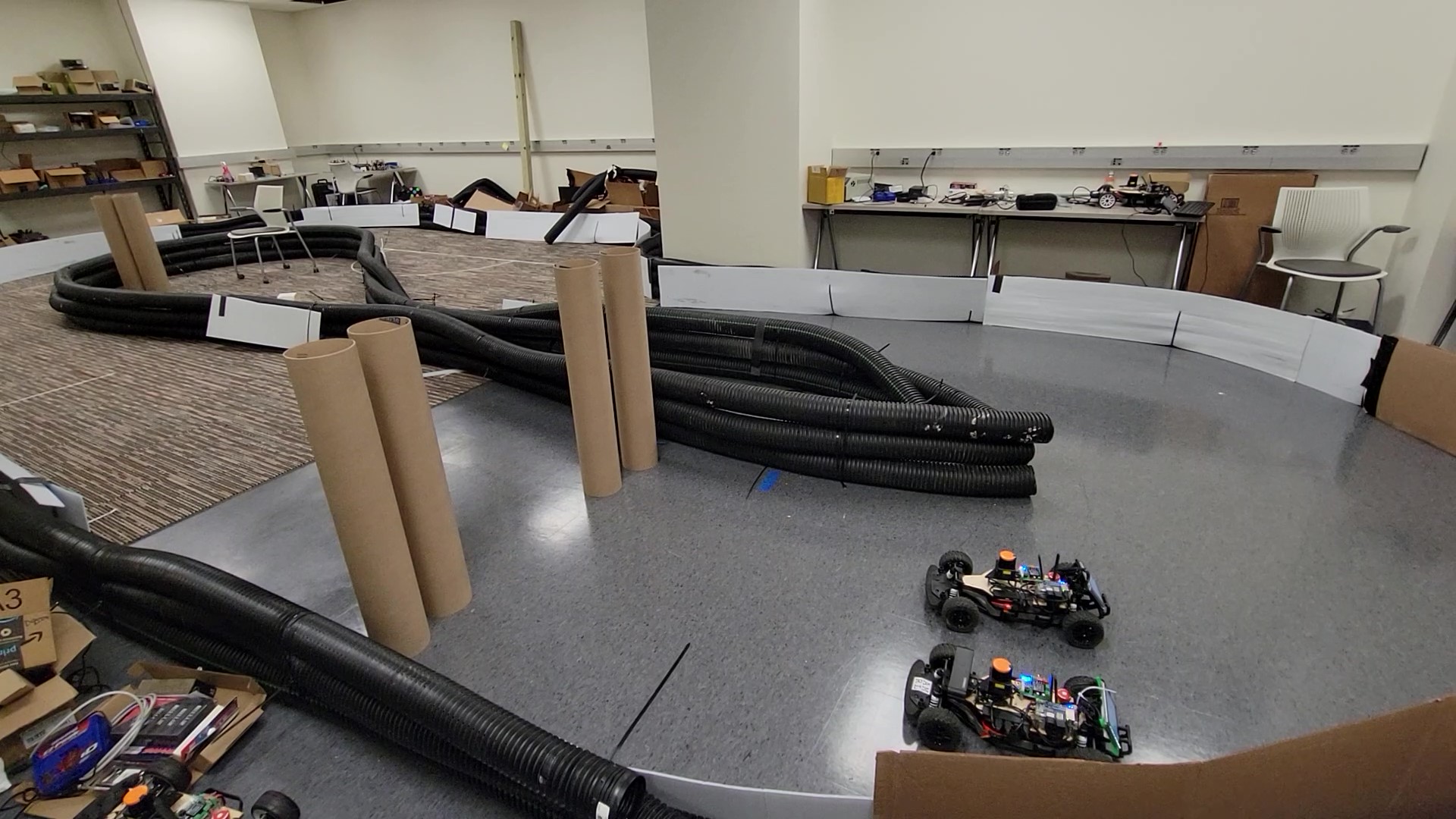}   
    \caption{\textit{No} scheduling, initial time.}
    \label{fig: b1}
    \end{subfigure}
    \begin{subfigure}[h]{0.485\columnwidth}
\includegraphics[width = \columnwidth]{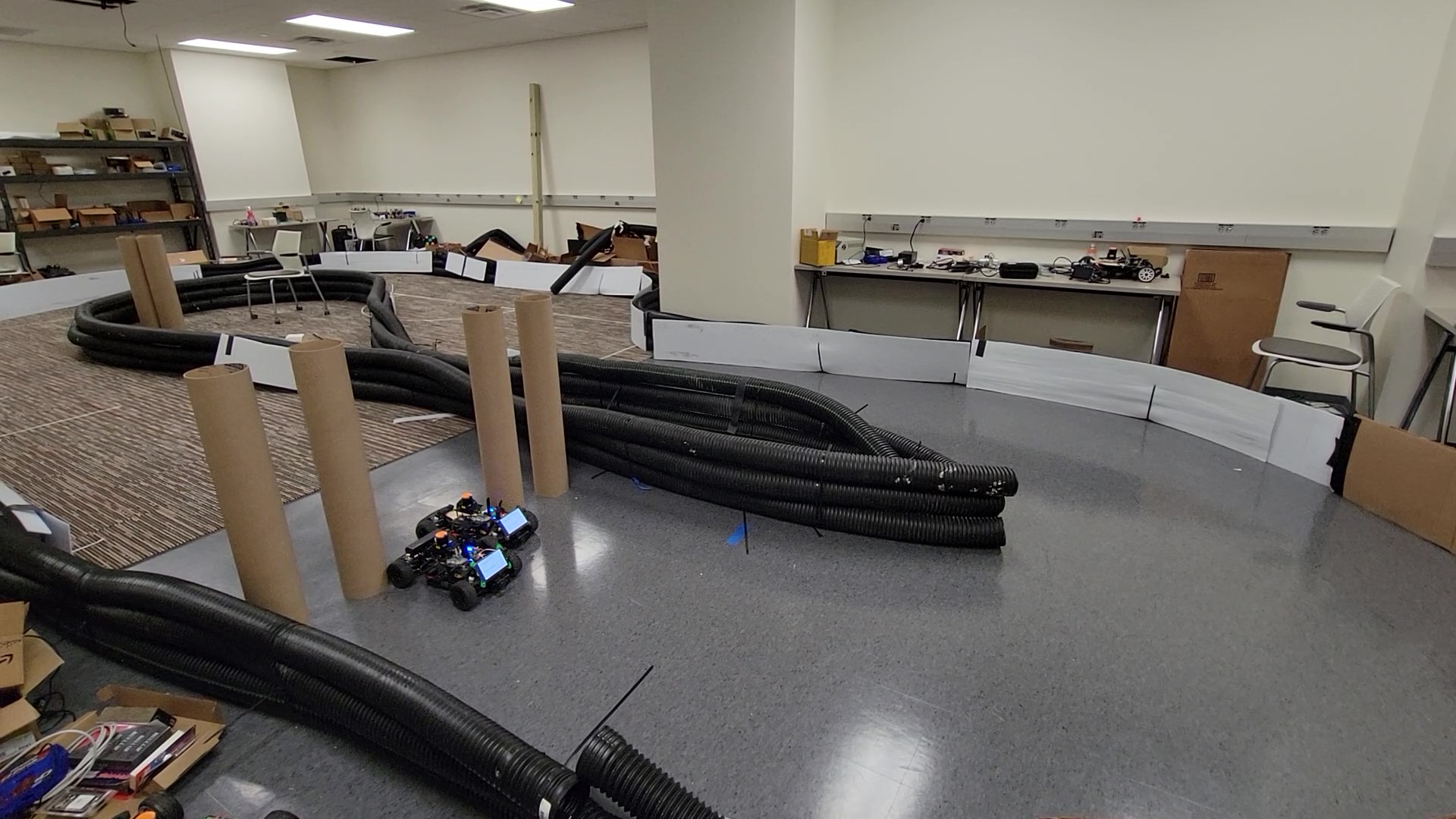}   
    \caption{\textit{No} schedule results in collision.}
    \label{fig: b2}
    \end{subfigure}
     \begin{subfigure}[h]{0.485\columnwidth}
\includegraphics[width = \columnwidth]{images/cover/ours1.jpg}   
    \caption{\textit{With} scheduling, initial time.}
    \label{fig: o1}
    \end{subfigure}
    \begin{subfigure}[h]{0.485\columnwidth}
\includegraphics[width = \columnwidth]{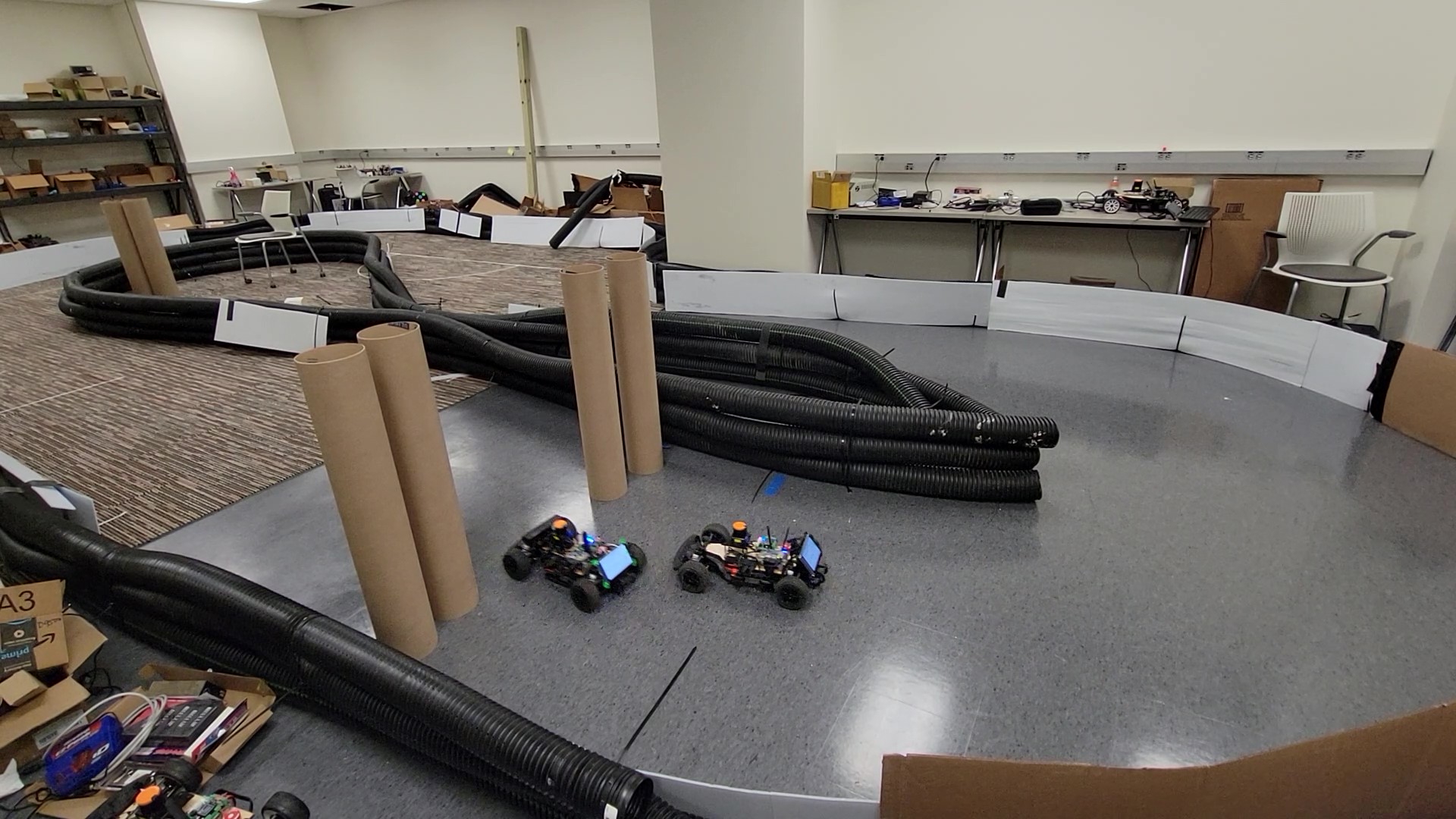}   
    \caption{Scheduling forms a queue.}
    \label{fig: o2}
    \end{subfigure}
    \caption{\textbf{Fully decentralized multi-agent social navigation:} \textit{(top)} Figures~\ref{fig: b1} and~\ref{fig: b2} depict two F$1/10$ cars at the initial time-step and when they collide at the gap. \textit{(bottom)} Figures~\ref{fig: o1} and~\ref{fig: o2} depict the same configurations but using our proposed bi-level optimization approach, both cars independently determine their priority order and form a queue to sfely navigate the doorway.}
    \label{fig: mini-games}
    \vspace{-10pt}
\end{figure}

A solution to addressing the limitations above may be found by drawing parallels with research methodologies in autonomous intersection management (AIM) in the traffic modeling literature~\cite{chandra_thesis,chandra2021meteor,chandra2022game, chandra2020forecasting, chandra2019traphic, suriyarachchi2022gameopt}. AIM is the task of managing vehicle traffic flow at unregulated traffic intersections in order to reduce congestion and improve travel times and fuel efficiency. AIM solutions employ a bi-level optimization strategy where the top-level optimization schedules the order of the vehicles through the intersection and the bottom level optimization solves an optimal trajectory planning problem constrained by the corresponding schedule. The main question addressed in this work is--\textit{can navigating social mini-games be modeled via a similar bi-level optimization strategy?} The top-level scheduling protocol requires a priority order that determines \textit{who} moves \textit{when} in the mini-game. Auctions are one such type of priority protocol that enables robots to decide their own priority in a decentralized fashion. Specifically, each robot bids whether to move or wait based on a specific bidding strategy. An auction program collects the bids and allocates each robot a turn that determines when they should move. Such a determination is meant to be acceptable to each robot. Furthermore, the robots make their bid without the knowledge of the objective functions of the other robots.

Determining the optimal priority order of moving through a doorway or a corridor intersection is, however, only half the solution. A robot must also be able to plan socially compliant trajectories that follow the determined ordering. A conservative, and indeed safe, planning output would be to simply emulate a ``stop sign'', where a robot would wait in place for a higher priority robot, moving only once the latter has passed through the doorway gap or intersection. Such a solution, while safe, is neither optimal, nor reflective of how humans navigate in similar scenarios. Crowd simulation and collision avoidance algorithms such as RVO~\cite{rvo} and Social Forces~\cite{social_forces} are more realistic but they do not model the dynamic constraints of robots and assume every robot cooperatively follows the same policy. Methods using deep reinforcement learning~\cite{mavrogiannis2020decentralized, wu2023iplan, sacadrl, long2018towards, ga3c-cadrl} train a navigation policy using gradient descent on simulated trajectory rollouts. But the policy is trained using a shared team reward and is identical across robots, again resulting in cooperative navigation. Moreover, these methods also suffer from the well-documented sim-to-real gap with poor success rates. Distributed optimization techniques~\cite{le2022algames, fridovich2020efficient, wang2021game} yield a game-theoretic solution to a constrained coupled optimization for non-cooperative robots where the objectives of other robots are assumed to be fully observable. Both reinforcement learning-based and optimization-based approaches, however, have not been successfully deployed on real robots due to sim2real gap or computational burden, or both.


\noindent\textbf{Main Contributions:} This paper presents a fully decentralized realtime bi-level optimization framework for socially compliant multi-robot navigation in social mini-games, such as navigating through a narrow door or negotiating right of way at a corridor intersection. The two levels of our bi-level optimization approach are:
    \begin{enumerate}
        \item \textit{Scheduling (top-level):} The top-level optimization determines a priority order for the robots to pass through a narrow gap or negotiate right of way at an intersection, based on robots' private incentives. Robots' participate in this scheduling protocol in a decentralized manner with the goal of maximizing their utility.
        \item \textit{Trajectory planning (bottom-level):} The bottom-level optimization is a motion planner that performs online greedy re-planning using a faster variant of the dynamic window approach~\cite{dwa}. Each robot only observes the positions and velocities of the other robots and computes the optimal trajectory in a locally distributed sense. The priority order from the scheduling protocol informs a linear velocity constraint on the robot's kinodynamic constraints.        
    \end{enumerate}

We successfully deploy the proposed algorithm in the real world using F$1/10$ robots, a Clearpath Jackal, and a Boston Dynamics Spot as well as in simulation using the SocialGym 2.0~\cite{socialgym} multi-agent social navigation simulator, in the doorway and corridor intersection scenarios. We compare with state-of-the-art social navigation methods using multi-agent reinforcement learning, collision avoidance algorithms, and crowd simulation models. Our results show that

\begin{itemize}
\item classical navigation performs $44\%$ better than the state-of-the-art learning-based social navigation algorithms.
\item without a scheduling protocol, our approach results in collisions in social mini-games
\item our approach yields $2\times$ and $5\times$ fewer velocity changes than CADRL in doorways and intersections.
\item bi-level navigation in doorways at a flow rate of $2.8 - 3.3$ (ms)$^{-1}$ is comparable to flow rate in human navigation at $4$ (ms)$^{-1}$.
\end{itemize}

\section{Related Work}
\label{ref: related_work}

In this section, we survey literature in different priority protocols and scheduling algorithms (Section~\ref{subsec: related_priority}), multi-agent trajectory planning (Section~\ref{subsec: related_trajectory_planning}), and navigation in different social contexts (Section~\ref{subsec: related_social_navigation})

\subsection{Priority Protocols and Scheduling Algorithms}
\label{subsec: related_priority}

Many different types of priority protocols and scheduling algorithms are used in the autonomous intersection management literature~\cite{aim_survey_1}. Some prominent protocols include first come first served (FCFS), auctions, and reservations.

FCFS~\cite{fcfs} assigns priorities to agents based on their arrival order at the intersection. 
It is easy to implement but can lead to long wait times and high congestion if multiple vehicles arrive at the intersection simultaneously. In auctions~\cite{carlino2013auction}, agents bid on the right to cross the intersection based on a specific bidding strategy. Auctions are more complex then FCFS, but allow agents to express their private priorities via decentralized bidding unlike FCFS, which is centralized. Reservation-based systems~\cite{reservation} are similar to the auction-based system in which agents reserve slots to cross the intersection based on their estimated arrival and clearance times. In this work, we employ the sponsored search auction~\cite{roughgarden2016twenty} to have robots decide their optimal priority order similar to prior wrok in the discrete setting~\cite{chandra2023socialmapf}.

\subsection{Multi-agent Trajectory Planning}
\label{subsec: related_trajectory_planning}

There is an extensive literature on multi-agent trajectory planning~\cite{motion_survey}. Relevant to this work includes general-sum differential games and deep reinforcement learning (DRL).

The literature on general-sum differential games classify existing algorithms for solving game-theoretic equilibria in robotics into four categories. First, there are algorithms based on decomposition~\cite{wang2021game, britzelmeier2019numerical}, such as Jacobi or Gauss-Siedel methods, that are easy to interpret and scale well with the number of players, but have slow convergence and may require many iterations to find a Nash equilibrium. The second category consists of algorithms based on dynamic programming~\cite{fisac2019hierarchical}, such as Markovian Stackelberg strategy, that capture the game-theoretic nature of problems but suffer from the curse of dimensionality and are limited to two players. The third category consists of algorithms based on differential dynamic programming~\cite{schwarting2021stochastic,sun2015game,sun2016stochastic,morimoto2003minimax,fridovich2020efficient,di2018differential}, such as robust control, that scale polynomially with the number of players and run in real-time, but do not handle constraints well. Lastly, the fourth category contains algorithms based on direct methods in trajectory optimization~\cite{le2022algames, di2019newton, di2020first}, such as Newton's method, that are capable of handling general state and control input constraints, and demonstrate fast convergence. 

DRL has been used to train navigation policies in simulation for multiple robots in social mini-games. Long et al~\cite{long2018towards} presents a DRL approach for multi-robot decentralized collision avoidance, using local sensory information. They present simulations in various group scenarios including social mini-games. CADRL~\cite{cadrl}, or Collision Avoidance with Deep Reinforcement Learning, is a state-of-the-art motion planning algorithm for social robot navigation using a sparse reward signal on reaching the goal and penalises robots for venturing close to other robots. A variant of CADRL uses LSTMs to select actions based on observations of a varying number of nearby robots~\cite{cadrl-lstm}. 

\subsection{Social Robot Navigation}
\label{subsec: related_social_navigation}

A survey on social navigation by Mavrogiannis et al.~\cite{social_survey_chris} highlights some of the initial efforts in the field~\cite{burgard1999experiences, thrun2000probabilistic, trautman2010unfreezing}. Since then, a flurry of research progress has been made including the development of social navigation simulators~\cite{biswas2022socnavbench, tsoi2022sean}. Researchers also focus on analyzing social cues and human walking behavior, such as gaits~\cite{gaits}, body language~\cite{body1}, and gaze~\cite{gaze_social}, to facilitate social robot navigation. For instance, gaze information can help robots predict human intention to avoid collisions. Body language can provide information about the presence of other people and their behavior, enabling robots to make informed decisions. Furthermore, there are also methods that attempt to model social norms such as right of way rules~\cite{sacadrl, chandra2020cmetric, chandra2021using, chandra2020graphrqi}.
\section{Preliminaries}

In this section, we begin by formulating a social mini-game navigation problem followed by stating the objective.

\subsection{Problem Formulation}
We describe a social mini-game as a partially observable stochastic game (POSG)~\cite{posg} with $k$ robots using the tuple, $\Big \langle k, \zeta^i, \mathcal{X},\{\mathcal{U}^i\},\{\mathcal{O}^i\},\{\Omega^i\}, \{\mathcal{T}^i\}, \{\mathcal{J}^i\}, \mathcal{R}, \mathcal{G} \Big\rangle$. Each robot is initialized with a start position $\left( x^i_I \right)$, a goal position $\left( x^i_G \right)$. The state space $\mathcal{X}$ is continuous; each robot has a pose $x^i_t\in \mathcal{X}$ and a private priority constant, $\zeta^i$. We are provided with a navigation graph $\mathcal{G}$ that is constructed using 2D vector graphics\footnote{\href{https://github.com/ut-amrl/vector\_display}{https://github.com/ut-amrl/vector\_display}}.

Further, each robot has a continuous action space $\mathcal{U}^i$ and an observation function $\mathcal{O}^i$ that takes in $x^i_t \in \mathcal{X}$ and outputs a local observation $o^i_t \in \Omega^i$ where $ o^i_t = [{x^i_t}, \tilde x^i_t]^\top$. Here, $\tilde x^i_t = \{x^j_t\}, j\in \mathcal{N}^i$, where $\mathcal{N}^i$ represents the robots close to robot $i$. Note that $o^i_t \in \Omega^i$ only contains observed pose of other robots and not their private incentives. The state transition function is given by $\mathcal{T}^{\Xi^i}:\mathcal{X}\times \mathcal{U} \longrightarrow \mathcal{X}$, where $\Xi^i$ represents the kinodynamic constraints for the $i^\textrm{th}$ robot. Each robot has a control policy $\pi^{\Xi^i}: \Omega^i \longrightarrow \mathcal{U}^i$ that takes in the local observation $o^i_t \in \Omega^i$ and performs a deterministic action $u^i_t \in \mathcal{U}^i $ to produce a trajectory $\Gamma^i = \Big( x^i_I, x^i_2, \ldots, x^i_G  \Big)$, where $x^i_{t+1} = g(x^i_t,o^i_t, u^i_t$ and $g(\cdot)$ is a low-level motion controller. The environment is geometrically constrained such that there exists a conflict zone $\varphi \in \mathcal{G}$.

Each robot has a cost function $\mathcal{J}^i:\Upsilon \longrightarrow \mathbb{R}$ that assigns a cost to a local trajectory according to various heuristic features such as obstacle clearance, length of the trajectory, distance from goal, and so on. Here, $\Gamma^i \in \Upsilon$ represents a trajectory in the set of all finite horizon trajectories $\Upsilon$. When an robot reaches its goal, it receives a utility, $\mathcal{\widehat J}^i(\Gamma^i = x^i_G) = \zeta^i\alpha_q$. Here, $\alpha_q\in \mathbb{R}$ belongs to a finite discrete set of distinct reward values, $\mathcal{R} = \{\alpha_1, \alpha_2,\ldots, \alpha_k\}$, where each robot is rewarded with a unique $\alpha_q$ indicate the time reward upon reaching on the $q^\textrm{th}$ turn. 

At this point, we are ready to state our problem objective.

\subsection{Problem Statement}

\begin{problem}
\textbf{Optimal Navigation in Social Mini-Games:} For the $i^\textrm{th}$ robot, an optimal solution to the social mini-game navigation problem is a collision-free trajectory $\Gamma^i_\textsc{opt}$ over the finite horizon $T$ such that $\Gamma^i_\textsc{opt} = x^i_I$ at $t=0$, $\Gamma^i_\textsc{opt} = x^i_G$ at $t=T$, $\Gamma^i_\textsc{opt} = \arg\min_{\Gamma^i \in \Upsilon} \mathcal{J}^i(\Gamma^i)$, and $\Gamma^i_\textsc{opt} = \arg\max_{\Gamma^i \in \Upsilon} \sum_i\mathcal{\widehat J}^i(\Gamma^i = x^i_G)$.
\label{prob: optimal_social_nav}
\end{problem}

A typical solution concept for non-cooperative POSGs is the \textit{Nash equilibrium}~\cite{gt-marl-survey} in which an optimal trajectory $\Gamma^i_\textsc{opt}$ is a best response to a given $\Gamma^{-i}_\textsc{opt}$, where $-i$ denotes every robot except the $i^\textrm{th}$ robot. That is, $\mathcal{J}^i(\Gamma^i_\textsc{opt}; \Gamma^{-i}_\textsc{opt}) \leq \mathcal{J}^i(\Gamma^i; \Gamma^{-i}_\textsc{opt}) \quad \forall \ i $. Informally, this indicates that a robot will have a best response strategy to a \textit{given} strategy profile of the other robots. A stronger solution is the \textit{(strictly) dominant strategy equilibrium} which states

\begin{equation}
    \mathcal{J}^i(\Gamma^i_\textsc{opt}; \Gamma^{-i}_\textsc{opt}) < \mathcal{J}^i(\Gamma^i; \Gamma^{-i}) \quad \forall \ i, -i
    \label{eq: dominant_strat}
\end{equation}

\noindent Informally, Equation~\ref{eq: dominant_strat} indicates that a robot will have a dominant strategy in response to any set of actions that other robots might take.

\section{Bi-Level Optimization Framework}

In order to solve Problem~\ref{prob: optimal_social_nav}, we propose a new bi-level 
optimization navigation framework of the following form,

\begin{equation}
    \Gamma^i_\textsc{opt} = \arg\min_{\Gamma^i \in \Upsilon^i} f^i\left(v^i, \mathcal{J}^i\right) \quad \forall i=1,2,\ldots,k
\vspace{-5pt}
\label{eq: snupi^individual_max}
\end{equation}

\noindent where $f^i(\cdot)$ is a non-linear function and $\Gamma^i_\textsc{opt}$ is a strictly dominant strategy solution. Key to this framework is the idea that in social mini-games, $f^i(\cdot)$ may be decomposed as $f^i(\cdot) = f^i_l(f^i_g(\cdot))$ where $f^i_g(\cdot)$ captures a \emph{global, discrete ordering} utility, while $f^i_l$ captures the \emph{local motion planning} cost. For example, the cost of a joint plan with two robots traversing through a narrow doorway can be decomposed into the utility of the \emph{order} in which the robots traverse through the door, and the cost of the motion trajectories that the robots execute while preserving such ordering. The outcome from maximizing $f^i_g(\cdot)$ is an optimal priority ordering, $\sigma_\textsc{opt} = \sigma_1\sigma_2\ldots\sigma_k$, over the robots where $\sigma^i =  j$ refers to the $i^\textrm{th}$ robot arriving at the goal on the $j^\textrm{th}$ turn.


The output of minimizing the cost of the local trajectory $\Gamma^i$ given $\sigma_\textsc{opt}$, $f_l(\cdot)$, is $\Gamma^i_\textsc{opt}$, which is the trajectory that enables the $i^\textrm{th}$ robot to reach the gap on turn $\sigma^i$. Represent the set of goal-reaching times for robot $i$ as $\Psi^i$. Further, let $\Psi^{\sigma^i} \subseteq \Psi^i$ be the set $\left \{t^i_g \ | \ t^{i^\prime}_g \leq t^i_g \leq t^{i^{\prime\prime}}_g\right\}$ that corresponds to the set of goal-reaching times such that robot $i$ arrives on turn $\sigma^i$, where $\sigma^{i^\prime} = \sigma^i - 1$ and $\sigma^{i^{\prime\prime}} = \sigma^i + 1$. Note that there must exist a set of kinodynamic controls, $\Xi^i_{\sigma^i}$, that allows the existence of $\Psi^{\sigma^i}$, which is the primary insight that we exploit in our implementation of the bi-level navigation framework described next.



\section{An Implementation of the Proposed Bi-Level Optimization Framework}
\label{sec: approach}

In this section, we present an implementation of the proposed bi-level optimization navigation framework described in the previous section. We first present an overview of our decentralized multi-robot navigation system. We describe each component of the system in detail below.
\begin{figure}[t]
    \centering
    \includegraphics[width=.85\columnwidth]{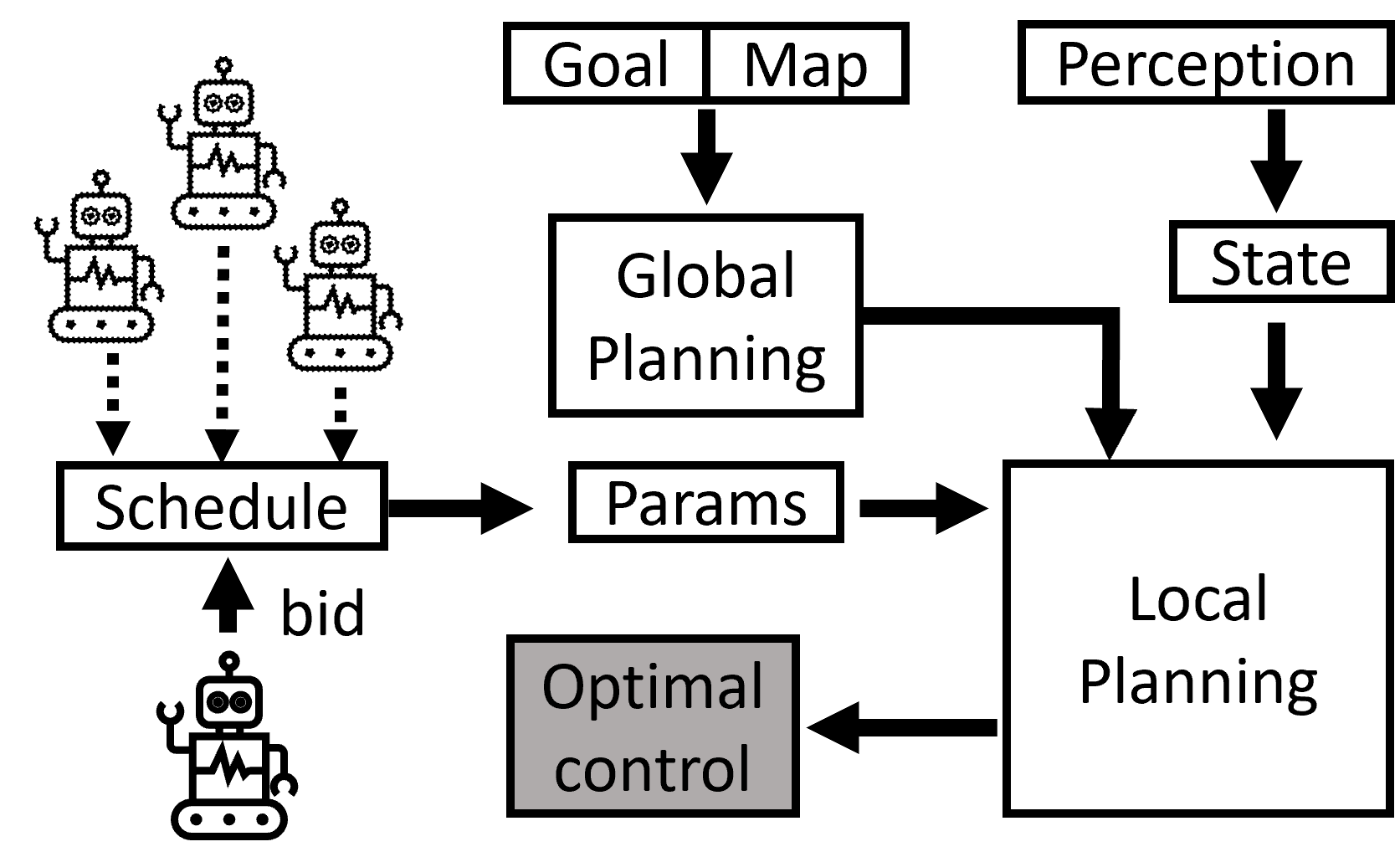}
    \caption{\textbf{Overview:} Our system comprises of a global planning phase followed by local planning, which includes kinodynamic control. The global planner takes as input a 2D vector map of the environment with static obstacles, \textit{e.g.}, walls, the current location of the robot, and the coordinates of a desired destination (or goal) in the map. Based on these inputs, the global planner computes a path in the map for the robot to reach the desired goal location. The local planner generates a feasible search in $\mathcal{X}$ over a short finite time horizon and computes the optimal trajectory for the robot to follow. The velocity controller computes the linear and angular velocities to steer the robot along the optimal trajectory. To achieve queue-formation in social mini-games, we implement an auction that uses robots' bids to compute an optimal priority order $\sigma_\textsc{opt}$ over the robots. Given $\sigma_\textsc{opt}$, we use hand-crafted rules for adding constraints on the kinodynamic controls for each robot such that the final trajectories follow $\sigma_\textsc{opt}$.
}
    \label{fig: overview}
    \vspace{-10pt}
\end{figure}
\subsection{Overview}
Each robot runs an instance of the following classical navigation framework. Our system comprises of a global planning phase followed by local planning, which includes kinodynamic control. We visualize the system in Figure~\ref{fig: overview}. The global planner takes as input a 2D vector map of the environment with static obstacles, \textit{e.g.}, walls, the current location of the robot, and the coordinates of a desired destination (or goal) in the map. Based on these inputs, the global planner computes an A$^*$ path in the map for the robot to reach the desired goal location. Robots determine the priority ordering in the social mini-game by each maximizing $f^i_g(v^i, r^i, p^i)$, where $f^i_g$ represents the utility from bidding in an incentive compatible auction with $r^i, p^i$ as the allocation and payment rules. Based on its valuation $v^i$, the robot will compute an optimal bid to submit to the auction independent of the valuations of all other robots. More specifically, $f^i_g(v^i, r^i, p^i)$ does not depend on the knowledge of $v^{-i}$.

The local planner is implemented using the University of Texas, Graph Navigation Stack, or simply \textit{GraphNav}\footnote{\href{https://github.com/ut-amrl/graph\_navigation}{https://github.com/ut-amrl/graph\_navigation}.}, which is a faster variant of the dynamic window approach (DWA)~\cite{dwa}. GraphNav uses the pure pursuit algorithm to determine local waypoints and generates a feasible search space in $\mathcal{X}$ over a short finite time horizon and computes the optimal trajectory for the robot to follow from that search space. Based on the ordering received from maximizing $f^i_g(\cdot)$, and features such as obstacle clearance and trajectory length, GraphNav generates $\Gamma^i_\textsc{opt} = \arg\min_{\Gamma^i \in \Upsilon}f^i_l(\mathcal{J}^i(\Gamma^i))$. Finally, the planner computes the linear and angular velocities to steer the robot along the optimal trajectory via a $1$D search. 

\subsection{Global Planning}

The global planning stage forms the first step in our navigation system. Given the 2D vector map representation of the environment and a corresponding a navigation graph, along with the start and goal locations, we compute a collision-free path (as a vector of graph vertices) from the initial location of the robot to the desired goal location using the A$^*$ algorithm. The global plan can be recomputed if necessary in each planning cycle.
\subsection{Priority Protocol}
In social mini-games, we define a conflict zone as a region $\varphi$ in the global map that overlaps goals corresponding to multiple robots. A conflict, then, is defined by the tuple $\langle \mathcal{C}^t_{\varphi}, \varphi, t \rangle$, which denotes a conflict between robots belonging to the set $\mathcal{C}^t_{\varphi}$ at time $t$ in the conflict zone $\varphi$ in $\mathcal{G}$. Naturally, robots must either find an alternate non-conflicting path or must move through $\varphi$ according to a schedule informed by a priority order. A priority ordering is defined as follows,

\begin{definition}
\textbf{Priority Orderings ($\sigma$): } A priority ordering is a permutation $\sigma:\mathcal{C}^t_{\varphi} \rightarrow [1,k]$ over the set $\mathcal{C}^t_{\varphi}$. For any $i,j \in [1,k]$, $\sigma^i = j$ indicates that the $i^\textrm{th}$ robot will move on the $j^\textrm{th}$ turn with $\sigma^{-1}(j) = i$.
\label{def: turn_based_ordering}
\end{definition}

\noindent For a given conflict $\langle \mathcal{C}^t_{\varphi}, \varphi, t \rangle$, there are $\left\lvert \mathcal{C}^t_{\varphi} \right\rvert$ factorial different permutations.
There exists, however, an \textit{optimal} priority ordering, $\sigma_\textsc{opt}$.

\begin{definition}
\textbf{Optimal Priority Ordering ($\sigma_{\textsc{opt}}$): } A priority ordering, $\sigma$, over a given set $\mathcal{C}^t_{\varphi}$ is optimal if bidding $b^i = \zeta^i$ is a dominant strategy and maximizes $\sum_{\lvert \mathcal{C}^t_{\varphi} \rvert} \zeta^i \alpha^i$.
\label{def: optimal_turn_based_ordering}
\end{definition}

\begin{algorithm}[t]
    \SetKwInOut{Input}{Input}
    \SetKwInOut{Output}{Output}
\SetKwComment{Comment}{$\triangleright$\ }{$40$Hz cycle frequency}
\SetAlgoLined
\Input{Bids $b^i$ for each agent in $\mathcal{C}^t_{\mathcal{O}}$, initialize $q \gets 0$.}
\Output{$\sigma_\textsc{opt}$}
     Sort $\mathcal{C}^t_{\mathcal{O}}$ in decreasing order of $b^i$.\\
     \While{$\lvert \mathcal{C}^t_{\mathcal{O}} \rvert$ is non-empty:}{
         Increment $q$ by $1$.\\
         Let $a^i$ be the the first element in $\mathcal{C}^t_{\mathcal{O}}$.\\
         Set $\sigma^i = q$.\\
     $\mathcal{C}^t_{\mathcal{O}} \gets \mathcal{C}^t_{\mathcal{O}} \setminus \{a^i\}$.\\
    }
     return $\sigma_\textsc{opt} = \sigma^1\sigma^2 \ldots \sigma_{\lvert \mathcal{C}^t_{\mathcal{O}} \rvert}$\\
\caption{Auction algorithm}
\label{algo: auction}
\end{algorithm}

We run an auction, $(r^i, p^i)$, with an allocation rule $r^i(b^i) = \sigma^i = q$ defined in Algorithm~\ref{algo: auction} and payment rule $p^i$ defined by $p^i(b^i) =  \sum_{j=q}^{\lvert \mathcal{C}^t_{\varphi} \rvert} \widehat b^{\sigma^{-1}(j+1)} \left( \alpha_j - \alpha_{j+1} \right)$. The payment rule is the ``social cost'' of reaching the goal ahead of the robots arriving on turns $q+1, q+2, \ldots, q+\mathcal{C}^t_{\varphi}$. The bids, $\widehat b^{\sigma^{-1}(q+1)}, \widehat b^{\sigma^{-1}(q+2)}, \ldots, \widehat b^{\sigma^{-1}\left(q+\lvert \mathcal{C}^t_{\varphi} \rvert\right)}$ represent proxy robot bids sampled from a uniform distribution, since robots do not have access to the bids of other robots. Using $(r^i, p^i)$ defined as above, each robot solves

\begin{equation}
b^{i,*} = \arg\min_{b^i} r^i(b^i) - p^i(b^i)
\label{eq: utility_social}
\end{equation}

It is known~\cite{roughgarden2016twenty, chandra2022gameplan} that the auction defined by $(r^i, p^i)$ yields $b^{i,*} = \zeta^i$ and maximizes $\sum_{\lvert \mathcal{C}^t_{\varphi} \rvert} \zeta^i \alpha^i$. To summarize the algorithm, the robot with the highest bid is allocated the highest priority and is allowed to move first, followed by the second-highest bid, and so on. 







\subsection{Local Planning}

During the current planning cycle, given as input the local observation, $o^i_t$, corresponding to a robot $i$ at time $t$, the global A$^*$ path, as well as its kinodynamic constraints $\Xi^i_{\sigma^i}$, the local planner and computes the optimal local trajectory $\Gamma^i$ along with the controls $v, \omega$ needed to steer along $\Gamma^i$.

Most importantly, the local planner first modifies the robot's default kinodynamic constraints based on the value of $\sigma^i$. In this system, we employ a linear maximum velocity scaling of $\frac{\text{V\_MAX}}{\sigma^i}$. So for example, if $\sigma^i = 1$, then $\texttt{V\_MAX}$ for the first robot is unchanged, for the second robot it will become $\frac{\texttt{V\_MAX}}{2}$, and so on. We denote these modified kinodynamics by $\Xi^i_{\sigma^i}$. Note that $\texttt{V\_MAX}$ refers to the maximum speed of the first robot. Since the pose and velocities of all robots are assumed to be visible, all subsequent robots can observe $\texttt{V\_MAX}$, allowing them to modify their own speeds.

The local planner proceeds computes a local target using the pure pursuit algorithm. Pure pursuit constrains the local target to be a point on the global path. Depending on the radius chosen, we can balance the trade-off between following the global path versus the sequence of locally planned trajectories. Once the local target is computed, the planner compute the optimal $\Gamma^i$ and $v, \omega$ in a DWA-like fashion. First, we generate a set of circular arcs, $\Upsilon^i_c$ from the current pose of the robot corresponding to a curvature of radius $c$. The optimal arc is selected according to the following optimization,

\begin{equation}
\begin{split}
    \Gamma^i_\textsc{opt} &= \arg\min_{\Gamma^i \in \Upsilon^i_c} \mathcal{J}^i(\Gamma^i)\\
                          &= \arg\min_{\Gamma^i \in \Upsilon^i_c} \sum_m w_m\phi_m(\Gamma^i)\\
\end{split}
\end{equation}

where $\sum_m w_m\phi_m(\Gamma^i)$ is a weighted linear sum of $m$ different features such as distance from obstacles, length of the trajectory, distance from goal, etc., where $w_m \in \mathbb{R}$ is a non-negative weight value and $\phi_m \in \mathbb{R}$ is a $m^\textrm{th}$ feature value that encodes the \textit{quality} of the trajectory $\Gamma^i$.

The final step of the local planner during the current planning cycle is to determine the optimal linear and angular velocities that steer the robot along $\Gamma^i_\textsc{opt}$. The linear velocity controller is a simple 1-D program that chooses between cruising, accelerating, and braking, based on the current speed at time $t$ and based the kinodynamic parameters, $\Xi^i_{\sigma^i}$, including maximum linear speed and acceleration stopping distances, along with the distance to goal $d^i_g$.

\section{Simulated and Real World Experiments}
\begin{figure}[t]
    \centering
    \begin{subfigure}[h]{0.49\columnwidth}
\includegraphics[width = \columnwidth]{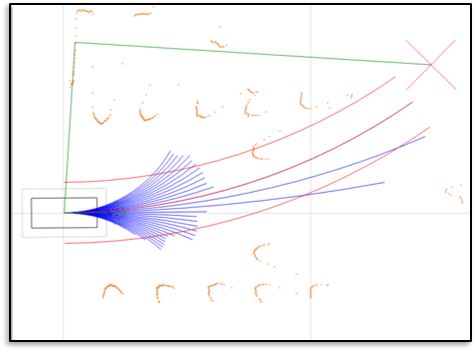}   
    \caption{Simple doorway environment.}
    \label{fig: long_traj}
    \end{subfigure}
    \begin{subfigure}[h]{0.49\columnwidth}
    \includegraphics[width=\columnwidth]{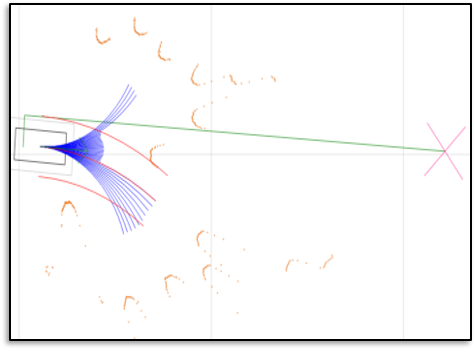}
    \caption{Static obstacles in robots path.}
    \label{fig: clipped_traj}
    \end{subfigure}
    \caption{\textbf{Visualizing the local planner:} The orange clusters represent lidar scans of the obstacles, the red cross is the local waypoint, the blue curves represent the 2D trajectory search space, and the red outer curves represent the minimum and maximum curvatures. In Figure~\ref{fig: long_traj}, we highlight that the optimal trajectory is one of the two longer blue curves that extend closer to the goal. In Figure~\ref{fig: clipped_traj}, we show that the search space gets clipped due to an obstacle in front.}
    \label{fig: mini-games}
    \vspace{-10pt}
\end{figure}
In this section, we present the outcomes of the bi-level optimization navigation approach in both simulation as well as on the UT Automata F$1/10$ car platforms, Boston Dynamics Spot, and the Clearpath Jackal in the doorway and intersection mini-games\footnote{our approach can, however, extend to the other mini-games.}. Through our experiments, we gain $4$ new insights that are important for the social robot navigation research community--$(i)$ classical navigation performs far better than learning-based algorithms for multi-agent social robot navigation in terms of success rate, $(ii)$ a scheduling protocol is necessary for distributed multi-robot social navigation, $(iii)$ our approach is more socially compliant, that is, results in the fewest changes in velocities by up to $5\times$, compared to the state-of-the-art social navigation method, CADRL~\cite{cadrl}, and $(iv)$ given a solution to the scheduling problem, there exists an optimal linear velocity constraint that enables collision-free multi-robot navigation.

\subsection{Experiment Setup}
\label{subsec: minigames}
\paragraph{Simulation Experiments}
\begin{figure}[t]
    \centering
    \begin{subfigure}[h]{0.49\columnwidth}
\includegraphics[width = \columnwidth, height=2.5cm]{images/cover/ours2.jpg}   
    \caption{Doorway}
    \label{fig: doorway_visual}
    \end{subfigure}
    \begin{subfigure}[h]{0.49\columnwidth}
    \includegraphics[width=\columnwidth, height=2.5cm]{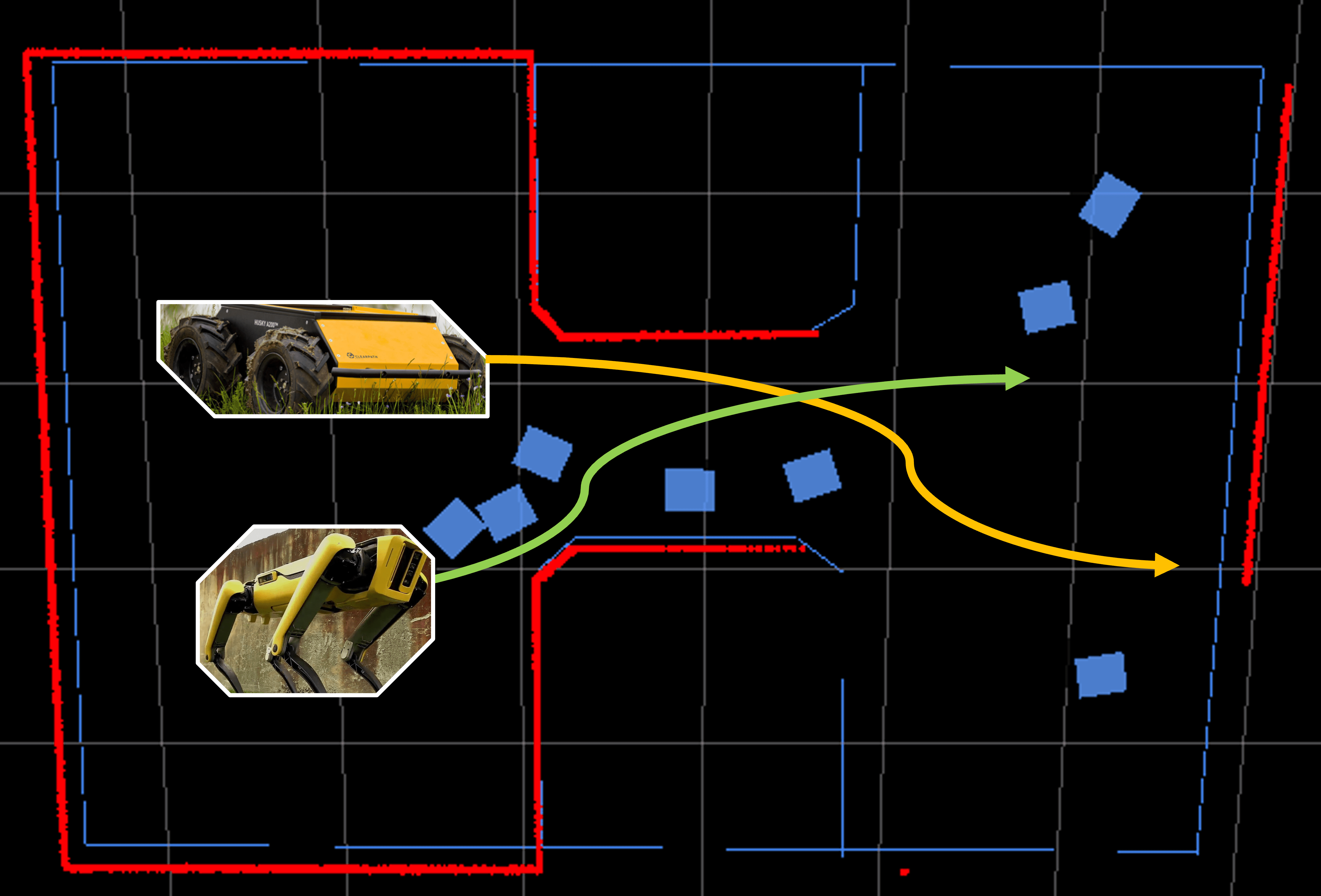}
    \caption{Doorway (simulation)}
    \label{fig: simulator}
    \end{subfigure}
    \begin{subfigure}[h]{\columnwidth}
\includegraphics[width = \columnwidth]{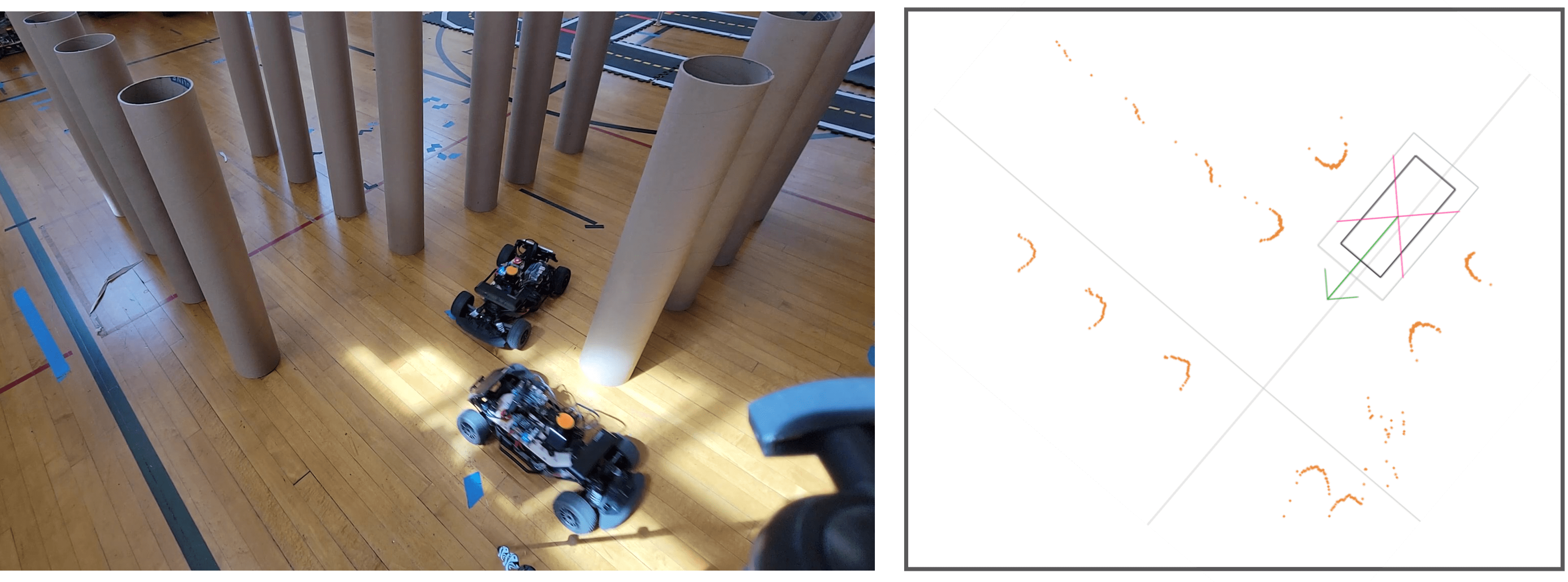}    
    \caption{Intersection and Visualizer}
    \label{fig: intersection_visual}
    \end{subfigure}
    \caption{\textbf{Visualizing and Simulating Social Mini-Games:} In Figures~\ref{fig: doorway_visual} and~\ref{fig: simulator}, we show an example of a doorway environment along with a simulation of the doorway in SocialGym 2.0~\cite{socialgym}. In Figure~\ref{fig: intersection_visual}, we show an example of the intersection environment along with its visualization in Webviz.}
    \label{fig: mini-games}
\end{figure}
We conduct simulation experiments using the SocialGym 2.0 multi-agent social navigation simulator, which is available on Github\footnote{\href{https://github.com/ut-amrl/social\_gym}{https://github.com/ut-amrl/social\_gym}} and described in detail in~\cite{socialgym}. To train the robots using multi-agent reinforcement learning, we use a Stable Baselines3 PPO policy with a step size of $4096$ and an MLP architecture. The training process consists of $1.25$ million steps, where the number of robots increases from $3$ to $5$ over time. Once trained, we evaluate the policies on $25$ trials for $3, 4, 5, 7$, and $10$ robot settings. During the training and evaluation phases, robots can perceive other robots' positions and velocities, their own distance to the goal, collision, and successes.

The reward function is designed to encourage robots to reach the goal efficiently while avoiding collisions. Robots receive a penalty of $-1$ for every step they are not at the goal, a reward of $100$ for reaching the goal, and a penalty of $-10$ for colliding with another robot. Additionally, robots receive a variable reward based on their progress towards the goal, calculated as the difference between their current location and their previous location. If no robots make significant progress towards the goal within $100$ steps (as measured by a total magnitude of $0.5$ meters), the episode ends and all robots not at the goal receive a penalty of $-100,000$ 

\paragraph{Real World Experiments}

We have designed and tested two distinct social mini-games, namely the \textit{doorway} and \textit{corridor intersection} scenarios, in a $3$ meters by $3$ meters space. Our testing involved up to $3$ robots, despite the fact that the space could accommodate up to $4$. In the doorway scenario, we set the gap size to approximately $0.5$ meters. In the corridor intersection scenario, we determined that the width of the four arms of the intersection should be between $1.5$ and $2$ meters, with the conflict zone at the center of the intersection having an area of $2.5$ to $4$ square meters. In the doorway scenario, all robots start from one side of the doorway, either equidistant from the gap at a distance of approximately $1.8$ meters or using staggered starting positions, and their objective is to move to the other side. In the corridor intersection scenario, we test standard autonomous intersection navigation with one robot on each arm of the intersection at any time. The goal here is to navigate the intersection safely. A robot on one arm of the intersection can select any of the 3 remaining arms to go to.

We test three distinct robots: the UT Automata F$1/10$ car platforms, the Clearpath Jackal, and the Boston Dynamics Spot. We selected these robots for their varying shape, size, and kinodynamic parameters. The Spot is a legged robot, while the Jackal and the F$1/10$ cars are wheeled. Additionally, the Jackal can make point turns, while the cars cannot. The Spot and the Jackal can move at maximum speeds of $1.5$ meters per second whereas the UT Automata F$1/10$ cars can travel up to $9$ meters per second. We experiment with various configurations with the exception of using the Spot and the Jackal with the F$1/10$ cars; the cars are not detected by the bigger robots on account of their low height.

\subsection{Bi-level Optimization versus Naive Multi-agent Planning on Real Robots}
\begin{figure}[t]
    \centering
    \begin{subfigure}[h]{0.325\columnwidth}
\includegraphics[width = \columnwidth]{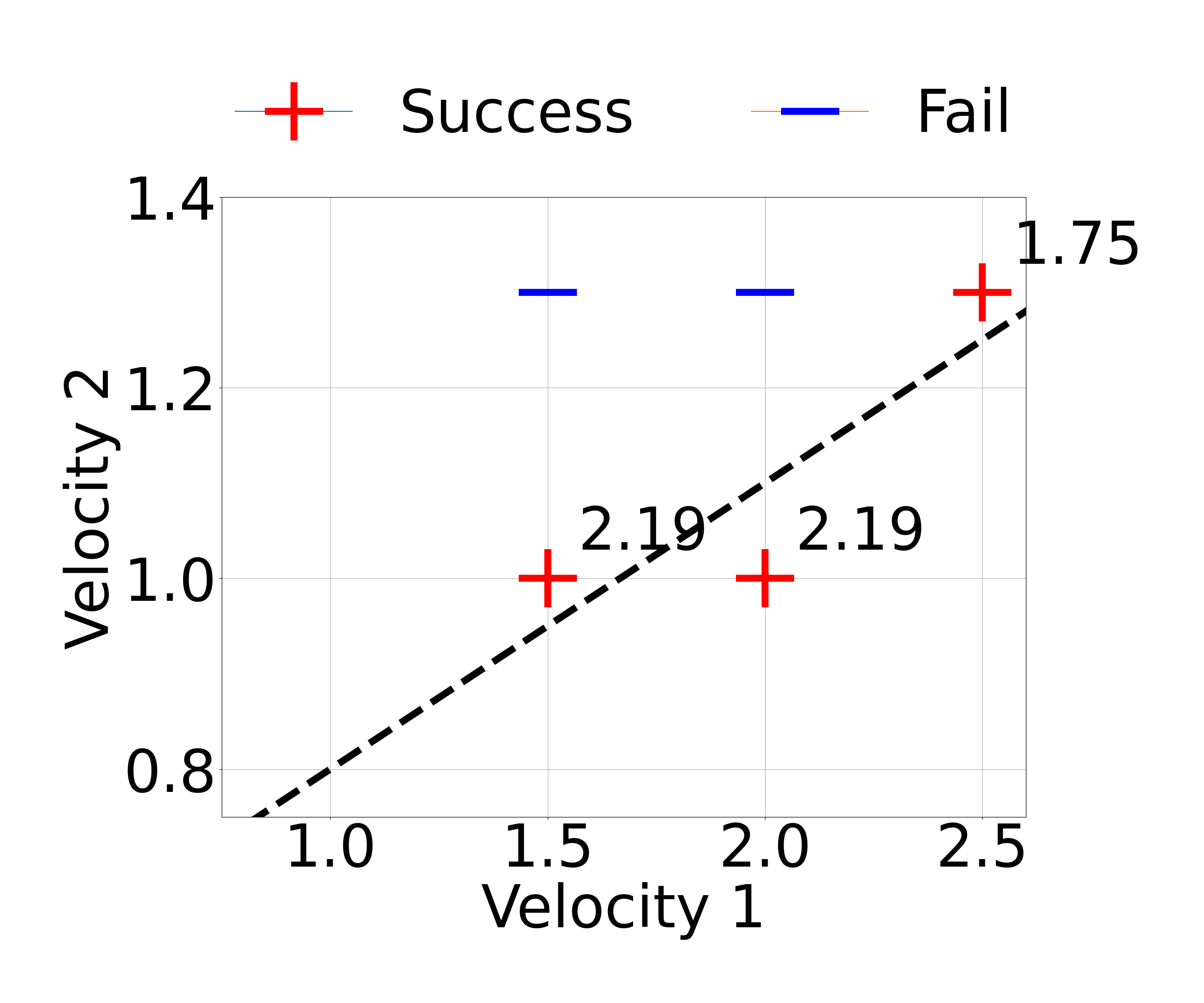}    
    \caption{F$1/10$, door.}
    \label{fig: doorway_result}
    \end{subfigure}
    \begin{subfigure}[h]{0.325\columnwidth}
\includegraphics[width = \columnwidth]{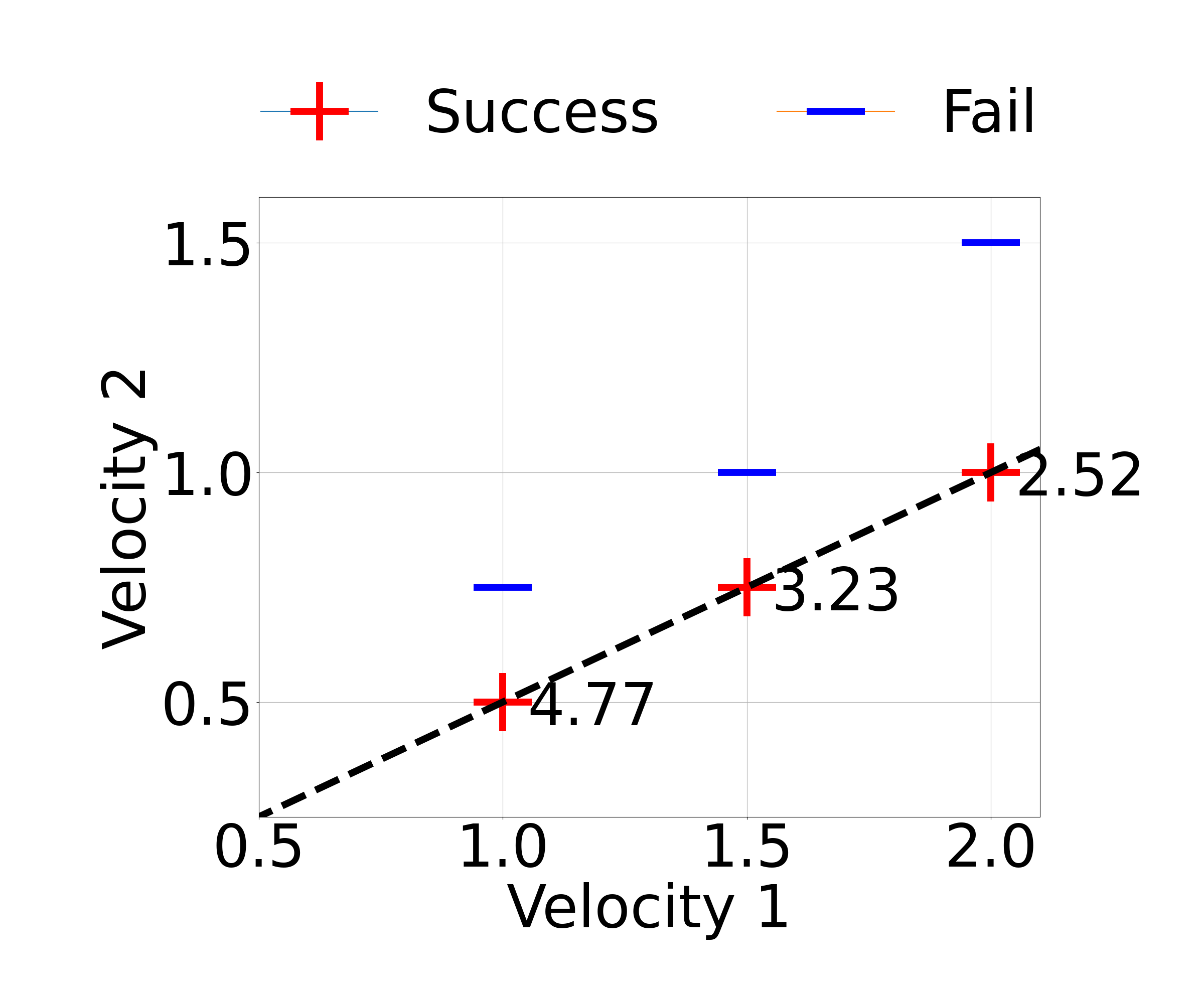}    
    \caption{F$1/10$, intersection.}
    \label{fig: intersection_result}
    \end{subfigure}
    \begin{subfigure}[h]{0.325\columnwidth}
\includegraphics[width = \columnwidth]{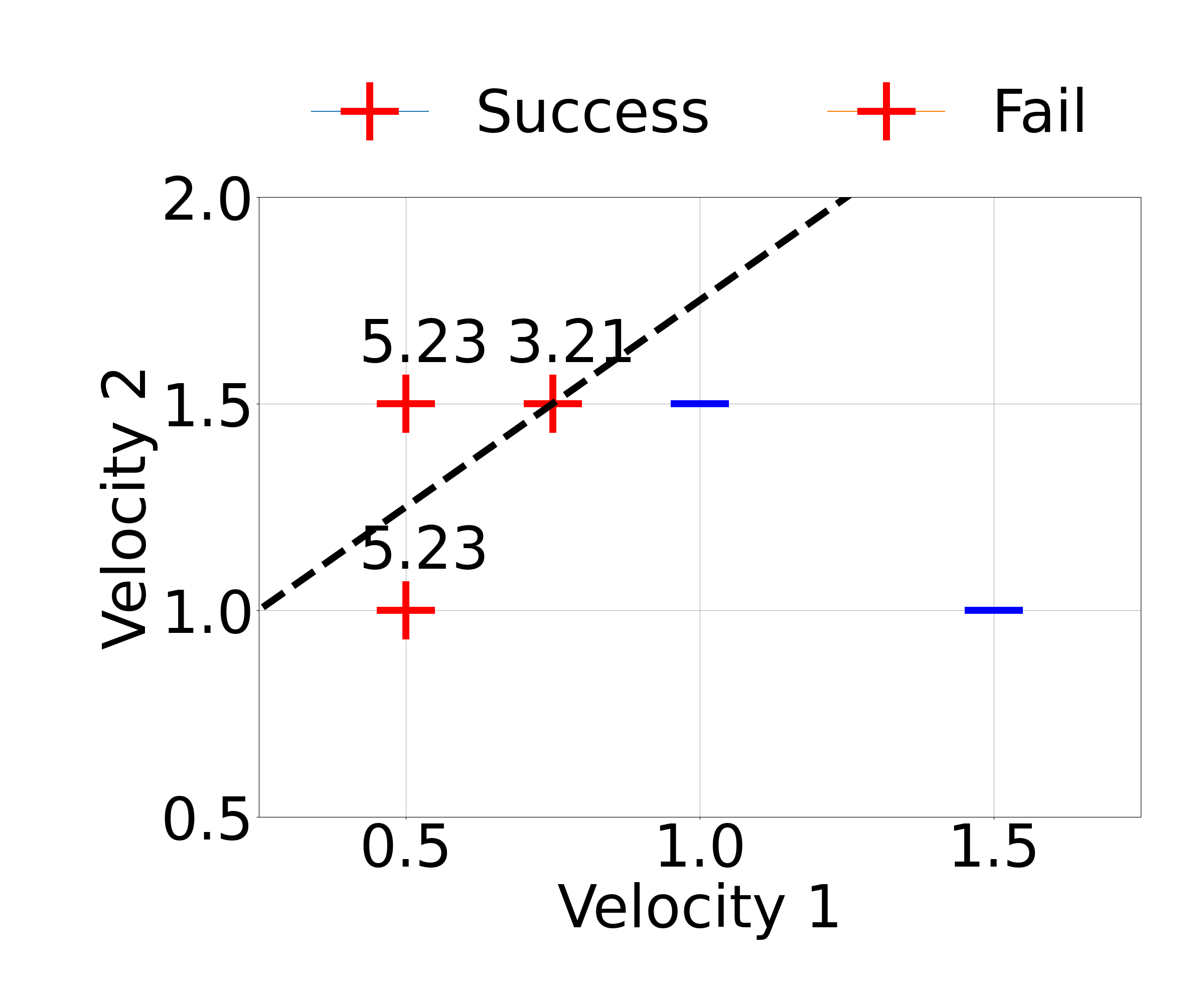}   
    \caption{Spot/Jackal, door.}
    \label{fig: jackal_result}
    \end{subfigure}
    \caption{\textbf{Results on $2$ F$1/10$ car platforms, Jackal, and the Spot robots:} We record successful and failed trials along with the makespan in the case of successful trials. Figures~\ref{fig: doorway_result} and~\ref{fig: intersection_result} use F$1/10$ cars while Figure~\ref{fig: jackal_result} shows results for the Jackal (``velocity 1'') versus the Spot (``velocity 2''). We notice that robots succeed when the velocity scaling is $\left({\sigma^i}\right)^{-1}$. }
    \label{fig: main_exps}
    \vspace{-10pt}
\end{figure}

We conducted experiments to evaluate the effectiveness of our bi-level optimization algorithm compared to a baseline that did not employ a priority ordering. The results are reported in Figure~\ref{fig: main_exps}. The baseline results are not reported as the robots failed $100\%$ of the time without a priority schedule. Here, we define a failure as not completing the mini-game due to collisions or deadlocks. Each graph in Figure~\ref{fig: main_exps} reports a successful or a failed trial, along with the makespan times (time to goal for the last robot to complete the mini-game) for each successful trial. 

We have found that our proposed bi-level navigation system is comparable to human navigation on average. According to several studies analyzing human walking speeds and makespan times at intersections~\cite{akcelik2001investigation} and doorways~\cite{garcimartin2016flow}, humans typically move at an average speed of $1.4$ meters per second at busy intersections at a flow rate of $4$ (ms)$^{-1}$\footnote{flow rate is measured in $\frac{N}{zT}$, where $N$ is the number of robots, $T$ is the makespan, and $z$ is the gap width in meters.}. When we compare this with our results in the doorway and intersection scenarios shown in Figures~\ref{fig: doorway_result} and \ref{fig: intersection_result}, we found that the robots moved at an average speed of approximately $1.25 - 1.5$ meters per second ($1.5/1.0$ and $2.0/1.0$ in Figure~\ref{fig: doorway_result}, $2.0/1.0$ in Figure~\ref{fig: intersection_result}) at a flow rate of $2.8 - 3.3$ (ms)$^{-1}$.

In Figure~\ref{fig: doorway_result}, we report the outcomes of trials conducted in the doorway setting using  $2$ F$1/10$ robots  We conducted $5$ trial runs, averaged across $3$ iterations, with varying velocities out which $3$ succeeded and $2$ failed. We observed that the trials that succeeded were where the cars' velocities were scaled approximately linearly by $1/\sigma^i$, whereas selecting velocities outside the $1/\sigma^i$ scaling resulted in collisions in the remaining $2$ trials. 

We repeated this experiment, this time with $6$ trials averaged across $3$ iterations, in the intersection scenario, where we again observed that the $2$ F$1/10$ cars succeeded in navigating the intersection with the  $1/\sigma^i$ scaling. We also conducted an experiment with $3$ F$1/10$ robots and used a staggered start configuration to reach the physical limit of feasible density. Both the doorway and intersection scenarios showed that our bi-level optimization algorithm improves the navigation system's performance and ensures safe trajectories. 
\begin{figure}[t]
    \centering
    \includegraphics[width=\columnwidth]{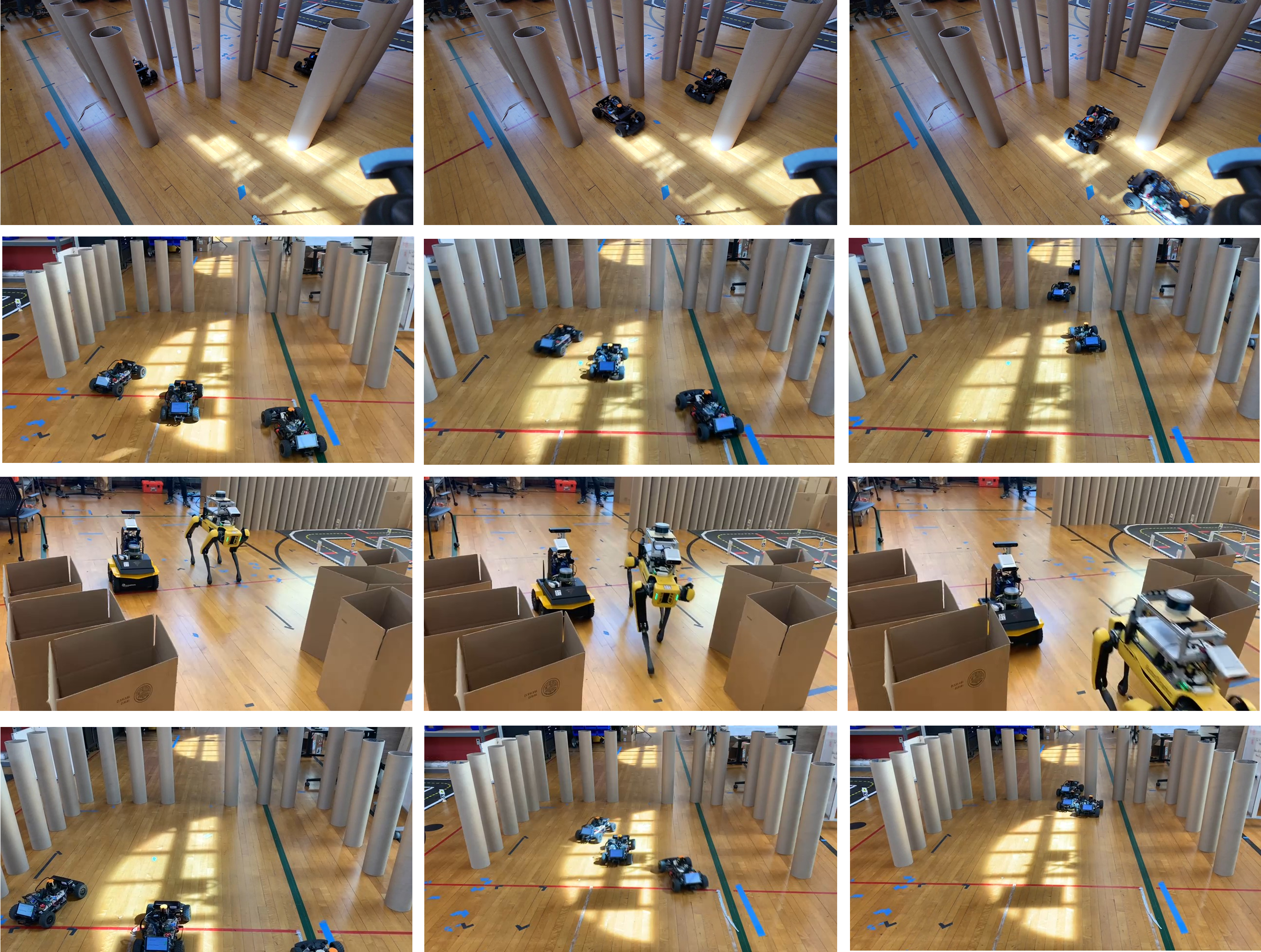}
    \caption{\textbf{Qualitative results on real robots:} \textit{(rows $1,2,3$)} We demonstrate our bi-level optimization approach at a corridor intersection and doorway using the F$1/10$ cars, Spot, and the Jackal robots. \textit{(row $4$)} We show decentralized local planning \textit{without} any scheduling protocol fails.  
    }
    \label{fig: qualitative}
    \vspace{-10pt}
\end{figure}

Finally, we tested the Spot and the Jackal robots in the doorway scenario using an identical setup as with the F$1/10$ platforms and report the results in Figure~\ref{fig: jackal_result}. Here also, we note the the emergence of the $1/\sigma^i$ scaling phenomenon. To gather more evidence for the benefits of this scaling, we scaled the velocities by $1/\sigma^i$ in the SocialGym 2.0 simulator and observed similar results. That is, our approach yields a $76\%$ and $96\%$ success rate with the $1/\sigma^i$ velocity scaling, but these rates drop to $48\%$ and $44\%$, respectively, when the scaling used is $\left (1-\frac{\sigma^i}{5\texttt{V\_MAX}}\right )^{-1}$. We present qualitative results of these experiments in Figure~\ref{fig: qualitative}.

\begin{table}[t]
    \centering
    \resizebox{\columnwidth}{!}{
    \begin{tabular}{crcccc}
         \toprule
          &Baseline& Success Rate &  Coll. Rate &Stop Time& Avg. $\Delta$V  \\
         \midrule
         \multirow{5}{*}{\rotatebox{90}{DOOR}}&CADRL~\cite{cadrl}&32&0.00&222&13 \\
        &CADRL(L)~\cite{cadrl-lstm}&0&0.12&436&36 \\
        &Enforced Order &44&0.16&617&117 \\
        &Only Local&4&2.32&166&30 \\
        &\cellcolor{green}Local w. priority&\cellcolor{green!25}76&0.17&\cellcolor{green!25}110&\cellcolor{green!25}6 \\
        \cmidrule{2-6}
       \multirow{5}{*}{\rotatebox{90}{INTER.}}&CADRL~\cite{cadrl}&20&1.20&389&28 \\
        &CADRL(L)~\cite{cadrl-lstm}&28&0.32&267&127 \\
        &Enforced Order &56&0.80&233&104 \\
        &Only Local&16&2.48&640&97 \\
        &\cellcolor{green}Local w. priority&\cellcolor{green!25}96&\cellcolor{green!25}0.14&\cellcolor{green!25}139&\cellcolor{green!25}6 \\ 
                
         \bottomrule
    \end{tabular}
    }
    \caption{\textbf{Benchmarking various MARL baselines:} We compare with $3$ MARL baselines and Only Local, which is an baseline method where only the local motion planner is used without scheduling. CADRL~\cite{cadrl} and its LSTM variant~\cite{cadrl-lstm} are state-of-the-art RL-based social navigation algorithms and Enforced Order uses sub-goal reward policies that encourage queue formation. Green indicates the best performing baseline for that scenario. \textbf{Conclusion:} Bi-level optimization results in successful and more socially compliant navigation compared to the state-of-the-art social navigation approach, CADRL.}
    \label{tab: main_exps_marl}
    \vspace{-10pt}
\end{table}

\subsection{Comparing with Learning-based Algorithms}

We benchmark $5$ multi-agent reinforcement learning (MARL) baselines--CADRL, CADRL(L), Enforced Order, and Only Local. CADRL~\cite{cadrl} and its LSTM-based variant, which we denote as CADRL(L), are state-of-the-art multi-agent social navigation methods. CADRL and CADRL(L) use a reward function where an robot is rewarded upon reaching the goal and penalized for getting too close or colliding with other robots as well as taking too long to reach the goal. We train these baselines using PettingZoo and Stable Baselines3~\cite{sb3} and report results across a range of social navigation metrics in Table~\ref{tab: main_exps_marl}. The metrics, each averaged across the number of episodes include success rate, time still, and average $\Delta$ velocity. The time still measures the number of times the robots had to stop and the average $\Delta$ velocity measures the numbers of times the robots had to deviate from their previous velocity.

Enforced Order is a MARL baseline, analogous to our approach, that encourages robots to engage in social behaviors such as queue formations. Finally, Only Local is an ablated baseline where we remove the high-level policy from the MARL interface, reducing it to a purely local multi-agent navigation baseline. We compare these baselines across several navigation metrics, following the standard literature and present results in Table~\ref{tab: main_exps_marl}.

We make $3$ observations. First, note that not only does our bi-level optimization approach outperform all other baselines (shown in green), but the second best performing baseline in the MARL category is Enforced Order ($44\%$ in doorway, $56\%$ in intersection), which also attempts to learn a type of priority ordering, signaling that determining an optimal schedule is indeed the key to success in multi-robot social navigation. Next, we observe that Only Local performs the worst--nearly failing in the doorway--which agrees with our real world experiments where the local navigation without the priority ordering failed $100\%$ of the time. Lastly, the average change in velocities with our approach is $2\times$ and $5\times$ smaller than the current best performing method, which is CADRL, indicating that our approach is more socially compliant that CADRL.

From some of the collision rates for the baselines reported in Table~\ref{tab: main_exps_marl}, we clarify that the the reason these ``collision-free'' MARL baselines yield low collision rates is because a majority of the episodes result in a deadlock and time out; the robots need not have collided. This can be verified by observing the corresponding abysmal success rates.

\subsection{Comparing with Crowd Simulation and Collision Avoidance Algorithms}
\begin{figure}[t]
\centering
   \begin{subfigure}[h]{0.49\columnwidth}
    \includegraphics[width=\textwidth]{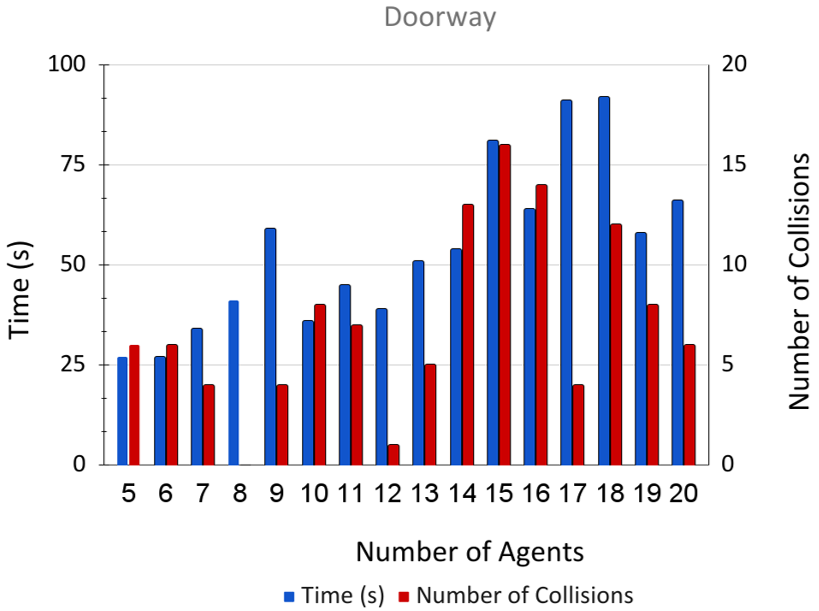}
    \caption{Reciprocal Velocity Obstacles}
    \label{fig: rvo1}
  \end{subfigure}
  %
  %
    %
   \begin{subfigure}[h]{0.49\columnwidth}
    \includegraphics[width=\textwidth]{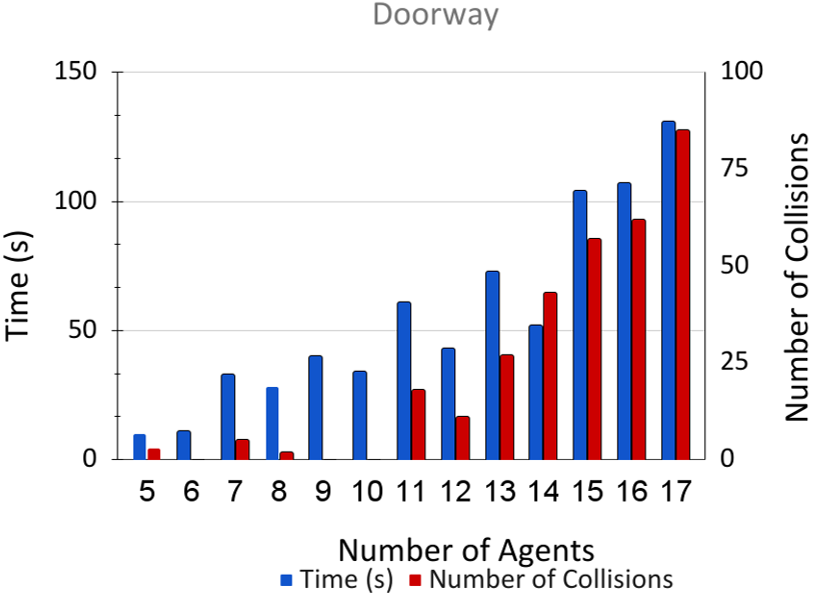}
    \caption{Social Forces}
    \label{fig: sf1}
  \end{subfigure}
  %
\caption{\textbf{Comparing with crowd simulation and collision avoidance models:} Measuring the time to goal and number of collisions with respect to number of agents for Reciprocal Velocity Obstacles (RVO)~\cite{rvo, orca} and Social Forces~\cite{socialforces}.}
  \label{fig: rvo_results}
  \vspace{-10pt}
\end{figure}  

Relevant to our evaluation and analysis are collision avoidance and crowd simulation models such as ORCA~\cite{orca} and social forces~\cite{social_forces} as they are also frequently used for social navigation models for cooperative robots. We test the capabilities of these models in the doorway and intersection scenarios by varying the number of robots and reporting the simulation time in seconds and the number of collisions in Figure~\ref{fig: rvo_results}. Figures~\ref{fig: rvo1} and~\ref{fig: sf1} show that the time to goal and the number of collisions generally increase with more robots. 

In RVO, most of the collisions occurred when the robots were approaching their goal. Because they all had the same goal, they would move around the goal close to each other and sometimes collide. However, there were fewer collisions in RVO than in social forces. In social forces, the robots would clump together at the goal (which would be considered a collision) so we disregarded whenever this happened when counting actual collisions. For Social Forces, most collisions occurred when the robots were approaching the narrow doorway. The robots would often oscillate back and forth near the doorway and end up colliding with another robot. All experiments involving social forces and RVO were performed using their default parameters with little to no tuning. It is possible that the number of collisions could be reduced by increasing the constant for repulsive forces or reducing attractive forces.

\section{Conclusion, Limitations, and Future Work}
This paper proposes a novel realtime fully decentralized approach to safe, efficient, and socially-compliant multi-robot navigation in social mini-games. The proposed approach uses a bi-level optimization framework that separates the determination of the priority order of the robots and the trajectory planning problem. The approach overcomes the limitations of current approaches based on deep reinforcement learning and crowd simulation models in real world settings. The proposed approach has been successfully deployed on the F$1/10$ car platforms, the Clearpath Jackal, and the Boston Dynamics Spot in navigating narrow doors, negotiating right of way at a corridor intersection, and other challenging scenarios.

Our approach has some limitations. Our navigation currently uses simple handcrafted rules to model agents' priorities, reducing its ability to generalize easily to novel environments. In the future, we will use inverse optimal control techniques to learn cost functions that implicitly embeds priority information for each agent.

\footnotesize{
\bibliography{refs}
\bibliographystyle{plain}
}

\end{document}